\documentclass[runningheads]{eecv_styles/llncs}

\usepackage[mobile]{eecv_styles/eccv}

\usepackage{eecv_styles/eccvabbrv}
\usepackage{graphicx}
\usepackage{amsmath}
\usepackage{tikz}
\usepackage{amssymb}
\usepackage{booktabs}
\usepackage[breaklinks,colorlinks]{hyperref}
\usepackage[utf8]{inputenc}

\usepackage{url}
\usepackage{multirow}
\usepackage{mathtools, nccmath}
\usepackage{amsfonts,dsfont,eucal}
\usepackage{listings}
\usepackage{xcolor}
\usepackage{subcaption}
\usepackage[square,sort,comma,numbers]{natbib}
\usepackage{orcidlink}

\DeclareMathOperator{\Tr}{Tr}
\newcommand{\schatten}[2]{\left\Vert #1 \right\Vert_{#2}}
\newcommand{\frobenius}[1]{\left\Vert #1 \right\Vert_F}
\newcommand{\frobeniussq}[1]{\left\Vert #1 \right\Vert_F^2}
\newcommand{\trace}[1]{\textrm{trace}(#1)}

\DeclareMathOperator*{\argmin}{arg\,min}
 \newcommand{\Hquad}{\hspace{0.5em}} 
 
\usepackage[capitalize]{cleveref}
\crefname{section}{Sec.}{Secs.}
\Crefname{section}{Section}{Sections}
\Crefname{table}{Table}{Tables}
\crefname{table}{Tab.}{Tabs.}

\usepackage[accsupp]{axessibility} 

\AtEndDocument{\refstepcounter{equation}\label{finaleq}}

\begin{document}

\title{FroSSL: Frobenius Norm Minimization for Efficient Multiview Self-Supervised Learning}

\author{Oscar Skean\inst{1}\orcidlink{0000-0002-4160-8392} \and
Aayush Dhakal\inst{2}\orcidlink{0000-0003-4431-0628} \and \\
Nathan Jacobs\inst{2}\orcidlink{0000-0002-4242-8967} \and
Luis Gonzalo Sanchez Giraldo\inst{1}\orcidlink{0000-0001-8984-9841}}

\authorrunning{O. Skean \etal}

\titlerunning{FroSSL}

\institute{University of Kentucky \and
Washington University in St. Louis \\ \email{oscar.skean@uky.edu}}

\maketitle
\begin{abstract}
    Self-supervised learning (SSL) is a  
    popular paradigm for representation learning. Recent multiview methods can be classified as sample-contrastive, dimension-contrastive, or asymmetric network-based, with each family having its own approach to avoiding informational collapse. While these families converge to solutions of similar quality, it can be empirically shown that some methods are epoch-inefficient and require longer training to reach a target performance. Two main approaches to improving efficiency are covariance eigenvalue regularization and using more views. However, these two approaches are difficult to combine due to the computational complexity of computing eigenvalues. We present the objective function FroSSL which reconciles both approaches while avoiding eigendecomposition entirely. FroSSL works by minimizing covariance Frobenius norms to avoid collapse and minimizing mean-squared error to achieve augmentation invariance. We show that FroSSL reaches competitive accuracies more quickly than any other SSL method and provide theoretical and empirical support that this faster convergence is due to how FroSSL affects the eigenvalues of the embedding covariance matrices. We also show that FroSSL learns competitive representations on linear probe evaluation when used to train a ResNet-18 on several datasets, including STL-10, Tiny ImageNet, and ImageNet-100. \href{https://github.com/OFSkean/FroSSL}{Github}
\end{abstract}

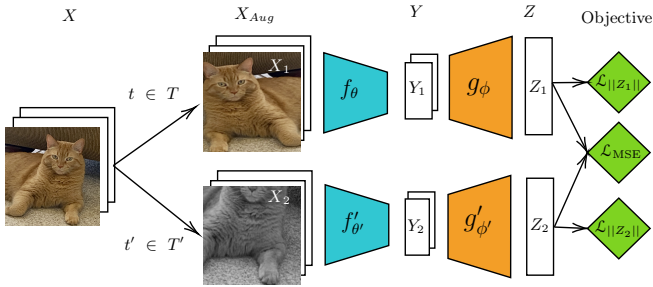
\begin{figure*}[!t]
\centering

\tikzset{every picture/.style={line width=0.75pt}} 

\scalebox{0.7}{\begin{tikzpicture}[x=0.75pt,y=0.75pt,yscale=-1,xscale=1]

\draw  [fill={rgb, 255:red, 255; green, 255; blue, 255 }  ,fill opacity=1 ] (294,138) -- (313.42,138) -- (313.42,178) -- (294,178) -- cycle ;
\draw  [fill={rgb, 255:red, 255; green, 255; blue, 255 }  ,fill opacity=1 ] (296,40) -- (315.42,40) -- (315.42,80) -- (296,80) -- cycle ;
\draw  [fill={rgb, 255:red, 255; green, 255; blue, 255 }  ,fill opacity=1 ] (156,123) -- (224.42,123) -- (224.42,190.25) -- (156,190.25) -- cycle ;
\draw  [fill={rgb, 255:red, 255; green, 255; blue, 255 }  ,fill opacity=1 ] (151,130) -- (219.42,130) -- (219.42,197.25) -- (151,197.25) -- cycle ;
\draw  [fill={rgb, 255:red, 255; green, 255; blue, 255 }  ,fill opacity=1 ] (155,27) -- (223.42,27) -- (223.42,94.25) -- (155,94.25) -- cycle ;
\draw  [fill={rgb, 255:red, 255; green, 255; blue, 255 }  ,fill opacity=1 ] (150,34) -- (218.42,34) -- (218.42,101.25) -- (150,101.25) -- cycle ;
\draw  [fill={rgb, 255:red, 255; green, 255; blue, 255 }  ,fill opacity=1 ] (13,78) -- (81.42,78) -- (81.42,145.25) -- (13,145.25) -- cycle ;
\draw  [fill={rgb, 255:red, 255; green, 255; blue, 255 }  ,fill opacity=1 ] (8,85) -- (76.42,85) -- (76.42,152.25) -- (8,152.25) -- cycle ;
\draw (181,74) node  {\includegraphics[width=52.5pt,height=52.5pt]{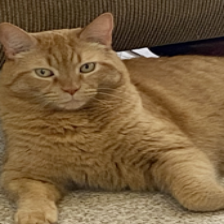}};
\draw (181,169) node  {\includegraphics[width=52.5pt,height=52.5pt]{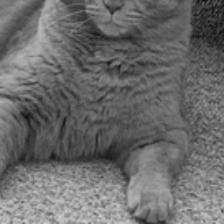}};
\draw  [fill={rgb, 255:red, 49; green, 199; blue, 210 }  ,fill opacity=1 ] (232.97,38.91) -- (279.03,52.73) -- (279.03,82.16) -- (232.97,95.98) -- cycle ;
\draw  [fill={rgb, 255:red, 49; green, 199; blue, 210 }  ,fill opacity=1 ] (233.97,132.91) -- (280.03,146.73) -- (280.03,176.16) -- (233.97,189.98) -- cycle ;
\draw  [fill={rgb, 255:red, 255; green, 255; blue, 255 }  ,fill opacity=1 ] (292,46) -- (311.42,46) -- (311.42,86) -- (292,86) -- cycle ;
\draw  [fill={rgb, 255:red, 255; green, 255; blue, 255 }  ,fill opacity=1 ] (292,143) -- (309.42,143) -- (309.42,183) -- (292,183) -- cycle ;
\draw (36.42,127) node [xscale=1] {\includegraphics[width=51pt,height=52.5pt]{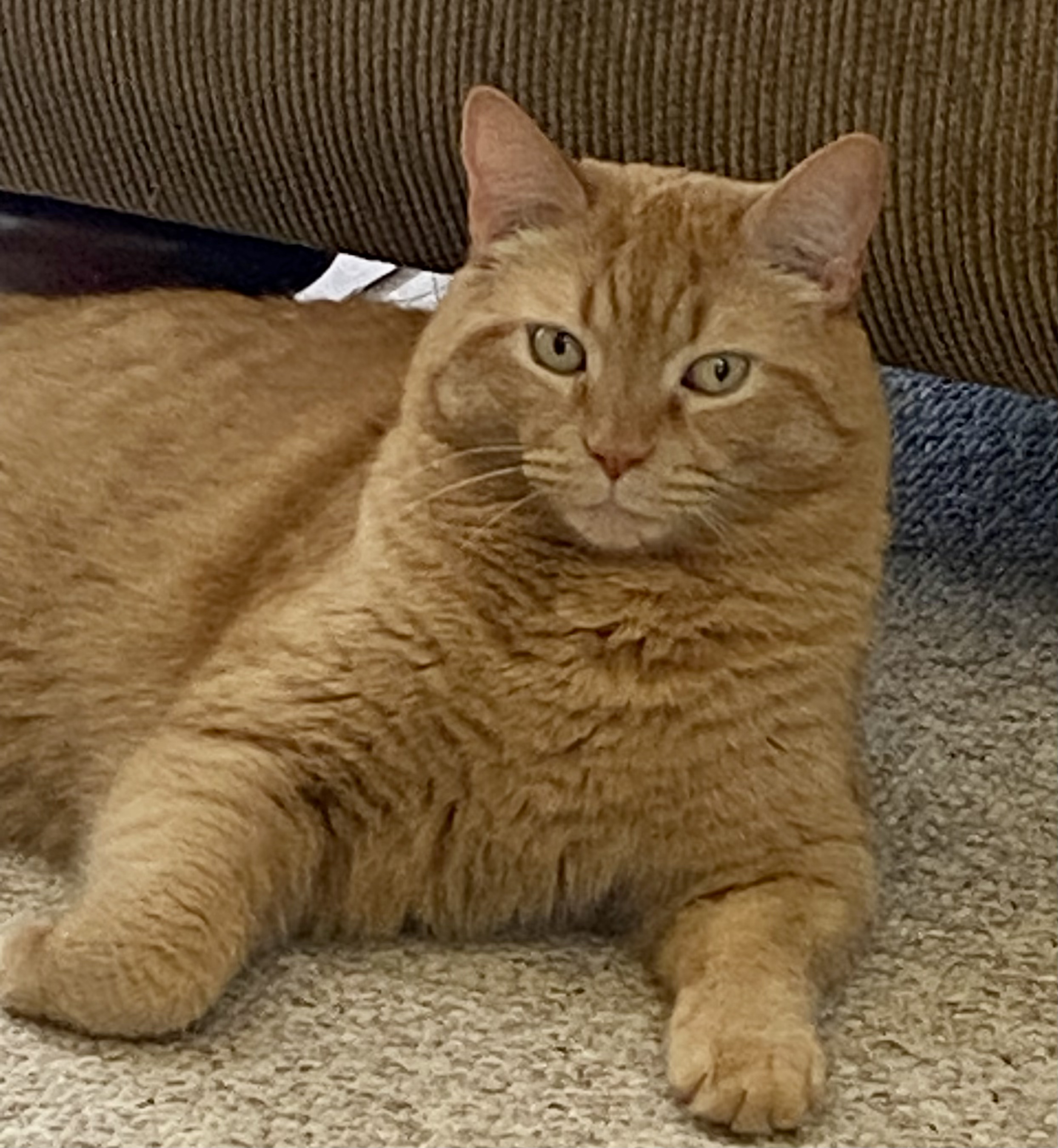}};
\draw    (80.42,119.25) -- (140.85,167.01) ;
\draw [shift={(142.42,168.25)}, rotate = 218.32] [color={rgb, 255:red, 0; green, 0; blue, 0 }  ][line width=0.75]    (10.93,-3.29) .. controls (6.95,-1.4) and (3.31,-0.3) .. (0,0) .. controls (3.31,0.3) and (6.95,1.4) .. (10.93,3.29)   ;
\draw    (80.42,119.25) -- (138.78,78.4) ;
\draw [shift={(140.42,77.25)}, rotate = 145.01] [color={rgb, 255:red, 0; green, 0; blue, 0 }  ][line width=0.75]    (10.93,-3.29) .. controls (6.95,-1.4) and (3.31,-0.3) .. (0,0) .. controls (3.31,0.3) and (6.95,1.4) .. (10.93,3.29)   ;
\draw  [fill={rgb, 255:red, 243; green, 158; blue, 39 }  ,fill opacity=1 ] (369.35,100) -- (323.95,86.33) -- (323.99,40.83) -- (369.42,27.25) -- cycle ;
\draw  [fill={rgb, 255:red, 243; green, 158; blue, 39 }  ,fill opacity=1 ] (368.35,199) -- (322.95,185.33) -- (322.99,139.83) -- (368.42,126.25) -- cycle ;
\draw  [fill={rgb, 255:red, 255; green, 255; blue, 255 }  ,fill opacity=1 ] (379,28.25) -- (398.42,28.25) -- (398.42,97.25) -- (379,97.25) -- cycle ;
\draw  [fill={rgb, 255:red, 255; green, 255; blue, 255 }  ,fill opacity=1 ] (380,128.25) -- (399.42,128.25) -- (399.42,197.25) -- (380,197.25) -- cycle ;
\draw  [fill={rgb, 255:red, 126; green, 211; blue, 33 }  ,fill opacity=1 ] (446.71,38.38) -- (469.42,60) -- (446.71,81.63) -- (424,60) -- cycle ;
\draw    (400,163) -- (422,163.92) ;
\draw [shift={(424,164)}, rotate = 182.39] [color={rgb, 255:red, 0; green, 0; blue, 0 }  ][line width=0.75]    (10.93,-3.29) .. controls (6.95,-1.4) and (3.31,-0.3) .. (0,0) .. controls (3.31,0.3) and (6.95,1.4) .. (10.93,3.29)   ;
\draw    (398.42,60.25) -- (422,60.02) ;
\draw [shift={(424,60)}, rotate = 179.44] [color={rgb, 255:red, 0; green, 0; blue, 0 }  ][line width=0.75]    (10.93,-3.29) .. controls (6.95,-1.4) and (3.31,-0.3) .. (0,0) .. controls (3.31,0.3) and (6.95,1.4) .. (10.93,3.29)   ;
\draw    (398.42,60.25) -- (423.09,108.22) ;
\draw [shift={(424,110)}, rotate = 242.79] [color={rgb, 255:red, 0; green, 0; blue, 0 }  ][line width=0.75]    (10.93,-3.29) .. controls (6.95,-1.4) and (3.31,-0.3) .. (0,0) .. controls (3.31,0.3) and (6.95,1.4) .. (10.93,3.29)   ;
\draw    (400,163) -- (423.17,111.82) ;
\draw [shift={(424,110)}, rotate = 114.36] [color={rgb, 255:red, 0; green, 0; blue, 0 }  ][line width=0.75]    (10.93,-3.29) .. controls (6.95,-1.4) and (3.31,-0.3) .. (0,0) .. controls (3.31,0.3) and (6.95,1.4) .. (10.93,3.29)   ;
\draw  [fill={rgb, 255:red, 126; green, 211; blue, 33 }  ,fill opacity=1 ] (446.71,88.38) -- (469.42,110) -- (446.71,131.63) -- (424,110) -- cycle ;
\draw  [fill={rgb, 255:red, 126; green, 211; blue, 33 }  ,fill opacity=1 ] (446.71,142.38) -- (469.42,164) -- (446.71,185.63) -- (424,164) -- cycle ;

\draw (292,1.4) node [anchor=north west][inner sep=0.75pt]    {$ \begin{array}{l}
Y\\
\end{array}$};
\draw (40,4.4) node [anchor=north west][inner sep=0.75pt]    {$ \begin{array}{l}
X\\
\end{array}$};
\draw (90,62.4) node [anchor=north west][inner sep=0.75pt]    {$t\ \in \ T$};
\draw (87,168.4) node [anchor=north west][inner sep=0.75pt]    {$t'\ \in \ T'$};
\draw (191,42) node [anchor=north west][inner sep=0.75pt]    {$\textcolor[rgb]{1,1,1}{X}\textcolor[rgb]{1,1,1}{_{1}}$};
\draw (191,135.4) node [anchor=north west][inner sep=0.75pt]    {$\textcolor[rgb]{1,1,1}{X}\textcolor[rgb]{1,1,1}{_{2}}$};
\draw (243,56.4) node [anchor=north west][inner sep=0.75pt]  [font=\large]  {$f_{\theta }$};
\draw (244,150.4) node [anchor=north west][inner sep=0.75pt]  [font=\large]  {$f'_{\theta '}$};
\draw (293,58) node [anchor=north west][inner sep=0.75pt]  [font=\footnotesize]  {$Y_1$};
\draw (292,156.4) node [anchor=north west][inner sep=0.75pt]  [font=\footnotesize]  {$Y_2$};
\draw (333,146.4) node [anchor=north west][inner sep=0.75pt]  [font=\Large]  {$g'_{\phi '}$};
\draw (335,55.4) node [anchor=north west][inner sep=0.75pt]  [font=\Large]  {$g_{\phi }$};
\draw (380,57.24) node [anchor=north west][inner sep=0.75pt]  [font=\footnotesize]  {$Z_1$};
\draw (381,155.24) node [anchor=north west][inner sep=0.75pt]  [font=\footnotesize]  {$Z_2$};
\draw (374,1.4) node [anchor=north west][inner sep=0.75pt]    {$ \begin{array}{l}
Z\\
\end{array}$};
\draw (428.5,52.4) node [anchor=north west][inner sep=0.75pt]  [font=\footnotesize]  {$\mathcal{L}_{||Z_1||}$};
\draw (428,156.4) node [anchor=north west][inner sep=0.75pt]  [font=\footnotesize]  {$\mathcal{L}_{||Z_2||}$};
\draw (432,102.4) node [anchor=north west][inner sep=0.75pt]  [font=\footnotesize]  {$\mathcal{L}_{\textrm{MSE}}$};
\draw (418,8) node [anchor=north west][inner sep=0.75pt]   [align=left] {Objective};
\draw (165,3.4) node [anchor=north west][inner sep=0.75pt]    {$ \begin{array}{l}
X_{Aug}{}\\
\end{array}$};

\end{tikzpicture}}

\caption{The SSL pipeline used in this work. In general, the encoder and projector may be asymmetric. We use symmetric encoders with shared weights and the same augmentation set for each view. We refer to $X_1$ as view 1 of $X$, and $X_2$ as view 2. Only two views are shown here, though more may be used in practice.}

\label{fig:ssl-diagram}
\end{figure*}

\section{Introduction}

The problem of learning representations without human supervision is fundamental in machine learning. Unsupervised representation learning is particularly useful when label information is difficult to obtain or noisy. It requires the identification of structure in data with limited knowledge about what the structure is.  One common way of learning structure without labels is joint embedding self-supervised learning (SSL)~\citep{chen2020simclr, haochen2021provable, tsai2021selfsupervised, chen2021simsiam, grill2020bootstrap, he2020momentum, zbontar2021barlow, li2021self}. The basic goal of SSL is to train neural networks to capture \textit{semantic} input features that are \textit{augmentation-invariant}. This goal is appealing for representation learning because the inference set often has similar semantic content to the training set.

A trivial solution to learning augmentation-invariant features is to encode all images to the same point, often called trivial or informational collapse. The resulting networks are essentially useless for downstream tasks. Different mechanisms have been proposed to handle collapse in SSL. These can be grouped into three families: sample-contrastive, dimension-contrastive, and asymmetric network methods. 

A less studied problem in all current SSL methods is their speed of convergence. When compared to traditional supervised learning, SSL methods must be trained for large numbers of iterations to reach competitive performance on downstream tasks. For example, a typical experiment in the literature is to train for 1000 epochs on ImageNet which can take several weeks even with many GPUs. An imperative direction of research is to investigate how to reduce SSL training time. An observation that is often hidden by only reporting the final epoch accuracy is that, empirically, certain SSL methods require more training time to reach competitive accuracies. This phenomenon has been observed for many dimension-contrastive methods by Simon \etal~\cite{simon2023stepwise} but not discussed in detail. We provide additional support for this claim in Section~\ref{sect:efficient-learning-eigenvalue}. Our work attempts to answer the following research question: Does there exist an SSL method with dimension-contrastive advantages, namely simplicity via avoidance of both negative sampling and architectural restrictions, while achieving competitive accuracies more quickly than other existing SSL methods?

We propose an SSL objective which we call FroSSL. Similar to many dimension-contrastive methods, FroSSL consists of a variance and invariance term. The invariance term is simply a mean-squared error between the views and is identical to VICReg's invariance term~\citep{bardes2022vicreg}. The variance term is the logarithm of the squared Frobenius norm of the normalized covariance embedding matrices. Using the Frobenius norm of covariance matrices for improving learned representations has not been explored in SSL. 

\newpage 

\noindent Our contribution can be summarized as:
\begin{itemize}
    
    \item We introduce the FroSSL objective function and show that it is \textit{both} dimension-contrastive and sample-contrastive up to a normalization of the embeddings.

    \item We introduce a theoretical framework that unifies dimension-contrastive methods that scale linearly in the number of views.
    
    \item We show that FroSSL combines two techniques to reduce training time: using more views and improving eigenvalue dynamics. We examine covariance eigenvalue trajectories during training on STL-10 to show that FroSSL learns useful, high-rank representations more quickly than other dimension-contrastive methods.
    
    
    \item We evaluate FroSSL on the standard setup of SSL pretraining and linear probe evaluation on CIFAR-10, CIFAR-100, STL-10, Tiny ImageNet, and ImageNet-100. We find that FroSSL achieves strong performance, especially when models are trained for fewer epochs.

\end{itemize}

\section{Background and Notation}
Consider a matrix $A \in \mathbb{R}^{m \times n}$. Let $A_{ij} \in \mathbb{R}$ denote the element at the $i$th row and $j$th column of $A$, and  $A_{i,:} \in \mathbb{R}^m$ denote the $i$th column vector representing the $i$th row of $A$, and $A_{:,j}$  the $j$th column of $A$. Let $\sigma_k(A)$ denote the $k$th singular value of $A$ ordered non-increasingly. The entry-wise product (also known as Hadamard product) is denoted as $A \odot B$. The Ky Fan $p$-norm of $A$ is defined as \cite{Horn2013matrix}:
\begin{equation} \label{eq:schatten_norm}
\displaystyle\schatten{A}{p} = \left(\sum_k^{\min(m,n)} \sigma_k^p(A)\right)^{1/p},
\end{equation}
which is a unitarily invariant norm. For $p = 2$, we have the Frobenius norm $\schatten{A}{2} = \frobenius{A} = \sqrt{\sum_{i} \sum_j A_{ij}^2}$.

\subsection{The Joint Embedding Self-Supervised Learning Problem}
\label{sect:ssl-problem}
The goal of self-supervised learning is to learn useful representations without external supervision. Many visual joint embedding SSL methods follow a similar procedure which was first introduced in \cite{chen2020simclr}. An example of this procedure is depicted in Figure~\ref{fig:ssl-diagram}. Let $\mathbf{X} = \{x_i\}_{i=1}^{N}$ be a minibatch with $N$ samples, $V$ the number of augmented views, $T(\cdot)$ a function that applies randomly selected augmentations to an image, $f$ a visual encoder network, and $g$ a projector network. 

First, each image $x_i \in \mathbf{X}$ is paired with augmented versions of itself, making the augmented dataset $\mathbf{X}_\textrm{aug} =\{T_1(x_i), \cdots, T_V(x_i)\}_{i=1}^{N} = \{X_1, \cdots, X_{V} \}$ With ideal augmentations, $\left(X_{1}\right)_{i, :}$ and $\left(X_{2}\right)_{i, :}$ have identical \textit{semantic} content and different \textit{style} content.  Note that typically $V=2$, but we make no such assumptions. For each augmentation, the embedding set is given by $Y_{v} = \left\{ f((X_{v})_{i, :}) \right\}_{i=1}^N$ and projection set  $Z_{v} = \left\{ g((Y_{v})_{i, :}) \right\}_{i=1}^N$. Finally, an SSL objective is computed on the projections and backpropagated through both networks. The goal of the objective is to ensure that encoded augmentations for the same image are mapped close together by the projector, i.e. $\left(Z_{a}\right)_{i,:}$ and $\left(Z_{b}\right)_{i,:}$ are close in some sense of distance for all $a, b = 1, 2, \dots, V$. At the same time,  projections should capture the variability among images. Thus the goal of SSL is to train the networks $f$ and $g$ to extract \textit{semantic} features that are invariant to any augmentations induced by $T(\cdot)$. In the following, we take a closer look at choices for the SSL objective.

\subsection{The Three Families of Joint Embedding SSL Objectives}
Objective functions for joint embedding self-supervised learning can be divided into three families. The first family consists of \textit{sample-contrastive} methods~\citep{chen2020simclr, haochen2021provable, tsai2021selfsupervised, he2020momentum, caron2020unsupervised} which use a contrastive loss to learn a representation that maps positive samples (augmentation of the same image) close together while pushing negative samples (different images) apart. These methods avoid collapse at the expense of making comparisons between positive and negative samples. 

The second family consists of \textit{asymmetric network} methods~\cite{chen2021simsiam, grill2020bootstrap, caron2021dino} which place restrictions on the architecture of the mapping network used, including asymmetrical encoders~\citep{chen2021simsiam, grill2020bootstrap},  momentum encoders~\citep{he2020momentum}, and stop gradients~\citep{chen2021simsiam, halvagal2023implicit}. While these methods can achieve great results, they are rooted in implementation details and there is no clear theoretical understanding of how they avoid collapse~\citep{bardes2022vicreg}.

The third, and most recent, family are the \textit{dimension-contrastive} methods, which are sometimes called negative-free contrastive~\citep{tsai2021note} or feature decorrelation methods~\citep{tao2022unigrad}. These methods operate by reducing the redundancy in feature dimensions.  Instead of examining \textit{where} samples live in feature space, these methods examine \textit{how} feature dimensions are being used. Methods in this family can avoid the use of negative samples while also not requiring restrictions in the network architecture to prevent collapse. Barlow Twins objective pushes the normalized cross-covariance between views towards the identity matrix~\citep{zbontar2021barlow}. VICReg consists of three terms: the invariance term enforces similarity in embeddings across views, while the variance/covariance terms regularize the covariance matrices of each view to prevent collapse~\citep{bardes2022vicreg}. W-MSE whitens and projects embeddings to the unit sphere before maximizing cosine similarity between positive samples~\citep{ermolov2021whitening}. I-VNE maximizes the von Neumann entropy of the embedding covariance matrices~\citep{kim2023vne}. Finally, CorInfoMax maximizes the $\log\det$ entropy of both views while minimizing mean-squared error~\citep{ozsoy2022self}.

\begin{table}[!t]
\centering
\caption{Taxonomy of dimension-contrastive SSL methods describing how they avoid informational collapse and achieve augmentation invariance in the $D_{\textrm{inv}}$ and $D_{\textrm{var}}$ framework of Section \ref{sect:dimension-contrastive}.}
\label{tab:dc-taxonomy}
 \renewcommand{\arraystretch}{1.3}
\scalebox{0.6}{
\begin{tabular}{llc}
\hline
Method        & Variance $D_{\textrm{var}}(\Sigma_{v}\Vert \mathbf{I})$                                                                                                                                                                                                             & \multicolumn{1}{l}{Invariance $D_{\textrm{inv}}(Z_{v}, Z_r)$}                                                                   \\ \hline
VICReg        & (Variance) Hinge loss on auto-covariance diagonal                                                                                                                                                                     & \multirow{5}{*}{\begin{tabular}[c]{@{}c@{}}MSE\\ \vspace{-1cm}$\frac{1}{N}\frobenius{Z_v - Z_r}^2$ \end{tabular}} \\
              & (Covariance) covariance off-diagonals per view                                                                                                                                                                        &                                                                                                  \\
              & $\sum\limits_{k}^{D} \max\left(0, 1 - \sqrt{\left(\Sigma_{v}\right)_{k,k} + \epsilon}\right) + \nu \frobenius{\Sigma_{v} - \Sigma_{v}\odot \mathbf{I}}^2$ &                                                                                                  \\ \cline{1-2}
W-MSE         & Implicit through whitening  that                                                                                                                                                                                         &                                                                                                  \\
              & $D_{\textrm{var}}(\Sigma_{v}\Vert \mathbf{I}) = 0$ since $\Sigma_{v} = \mathbf{I}$ for all $v$                                                                                                                                                                                                                      &                                                                                                  \\ \cline{1-2}
CorInfoMax    & Log Det Divergence: $D_{\log\det}(A\Vert B) = \trace{AB^{-1}} - D - \log\det(AB^{-1})$                                                                                                                                                                      &                                                                                                  \\
              &                      $D_{\log\det}(\Sigma_v + \epsilon \mathbf{I} \: \Vert\:  \mathbf{I}) = \trace{\Sigma_v + \epsilon \mathbf{I}} - D - \log\det(\Sigma_v + \epsilon \mathbf{I})$                                                                                                                                                                                                &                                                                                                  \\ \hline
I-VNE         & von Neumann Relative Entropy: $S_{1} (A \Vert B) = \trace{A \left(\ln{A} - \ln{B}\right)}$                                                                                                                                                                       & \multirow{2}{*}{\vspace{1mm} Cosine Similarity}                        \\
              &                                                                                           $S_1(\Sigma_{v}\Vert \mathbf{I}) =  \trace{\Sigma_v \ln{\Sigma_v}}$                                                                                                                                    &  \multicolumn{1}{r}{}                                                                             \\ \hline
FroSSL (ours) & $2$-Order Petz-Rényi Relative Entropy: $S_{2} (A\Vert B) = \log{\trace{A^{2} B^{-1}}}$                                                                                                                                                             & \multirow{1}{*}{\begin{tabular}[c]{@{}c@{}}MSE\\ \vspace{1mm}$\frac{1}{N}\frobenius{Z_v - Z_r}^2$ \end{tabular}}   \\
              &                                                                                           $S_2(\Sigma_{v}\Vert \mathbf{I}) = \ln{\left(\sum_{i=1}^N \sigma_i^2  \right)}= \ln \frobeniussq{\Sigma_v}$                                                                                                                                                                     &                                                                          \\ \hline
\end{tabular}}
\end{table}

\subsection{A Framework for Dimension-Contrastive Methods}
\label{sect:dimension-contrastive}

Many recent works in dimension-contrastive SSL, whether explicitly or implicitly, consist of a combination of two competing objectives: an augmentation \textbf{invariance} term that pulls different augmentations from the same image close together, and a \textbf{variance} term that avoids  collapse of the mapping by regulating variance. Below, we unify dimension-contrastive methods into a general framework that is parameterized by choices of two distances. By carefully selecting these distances, specific dimension-contrastive methods can be recovered.

Let $Z_{v} \in \mathbb{R}^{N \times D}$ be a batch of projections and $\Sigma_{v} = \frac{1}{N}\hat{Z}_{v}^T\hat{Z}_{v}$ the corresponding covariance, where $\hat{Z}_{v}$ are the centered projections. A dimension-contrastive objective can be written as follows:
\begin{equation}\label{eq:dimension-contrastive_objective}
  \min \sum\limits_{v=1}^{V-1}\sum\limits_{r=v+1}^{V}  D_{\textrm{inv}}(Z_{v}, Z_r) + \gamma \sum\limits_{v=1}^{V} D_{\textrm{var}}(\Sigma_{v} \Vert \mathbf{I}). 
\end{equation}
The first term of \eqref{eq:dimension-contrastive_objective} is the \textbf{invariance} term which minimizes the distance $D_{\textrm{inv}} : \mathbb{R}^{N \times D} \times \mathbb{R}^{N \times D} \mapsto \mathbb{R}_{\geq 0}$ between all pairs of augmentations. The second term of \eqref{eq:dimension-contrastive_objective} is a \textbf{variance} factor that forces the covariance of each augmentation to be close to identity according to a dissimilarity $D_{\textrm{var}}: \mathbb{R}^{D \times D} \times \mathbb{R}^{D \times D} \mapsto \mathbb{R}_{\geq0}$. For instance, in VICReg~\citep{bardes2022vicreg}, $D_{\textrm{inv}}(Z_{v}, Z_r) = \frobenius{Z_{v} - Z_r}^2$ and $D_{\textrm{var}}(\Sigma_{v}\Vert \mathbf{I}) = \sum\limits_{k}^{D} \max\left(0, 1 - \sqrt{\left(\Sigma_{v}\right)_{k,k} + \epsilon}\right) + \nu \frobenius{\Sigma_{v} - \Sigma_{v}\odot \mathbf{I}}^2$. Similarly, in CorInfoMax~\citep{ozsoy2022self} $D_{\textrm{inv}}(Z_{v}, Z_r)$ is the same as VICReg, but $D_{\textrm{var}}(\Sigma_{v}\Vert \mathbf{I})$ can be related to the $\log\det$ divergence $D_{\log\det}(A\Vert B) = \trace{AB^{-1}} - D - \log\det(AB^{-1})$ setting $A = \Sigma_{v} + \epsilon \mathbf{I}$ and $B$ to a scaling of identity due to the normalization step in their projector. In Table~\ref{tab:dc-taxonomy}, we show the dimension-contrastive methods which fit into this framework. We provide derivations in Appendix \ref{appendix:derivations}.
\subsubsection{Multiview Invariance Term}
In \eqref{eq:dimension-contrastive_objective} the invariance term requires $V(V-1)/2$
comparisons which scales quadratically with the number of views. However, if $D_{\textrm{inv}}(Z_{v}, Z_r) = \frobenius{Z_{v} - Z_r}^2$, then the invariance term may be simplified to 
\begin{equation}
    \label{eq:mean-target}
    \sum\limits_{v=1}^{V-1}\sum\limits_{r=v+1}^{V}  D_{\textrm{inv}}(Z_{v}, Z_r) = V\sum\limits_{v=1}^{V} D_{\textrm{inv}}\left(Z_{v}, \overline{Z} \right) 
\end{equation}
where $\overline{Z} = \frac{1}{V}\sum_{i=1}^V Z_i$ is the average projection across all views. If a method has a $D_{\textrm{inv}}$ that can be rewritten this way, we say the method scales linearly with views. All methods displayed in Table $\ref{tab:dc-taxonomy}$ have this property.

\begin{table}[!t]
\centering
\caption{Taxonomy of dimension-contrastive SSL methods showing which desirable criteria they fulfill.}
\label{tab:dc-criteria-taxonomy}
\setlength{\tabcolsep}{8pt}
\scalebox{0.8}{
\begin{tabular}{lcccc}
\hline
             & \begin{tabular}[c]{@{}l@{}}Invariant to\\ Projection\\ Rotations\end{tabular} & \begin{tabular}[c]{@{}l@{}}Manipulates\\ Eigenvalues\\ Explicitly\end{tabular} & \begin{tabular}[c]{@{}l@{}}Quadratic in\\ Batch Size and\\ Dimension\end{tabular} & \begin{tabular}[c]{@{}l@{}}Linear\\ in Views\end{tabular} \\ \hline
Barlow Twins & $\times$                                                                      & $\times$                                                                       & $\checkmark$                                                                     & $\times$                                                  \\
VICReg       & $\times$                                                                      & $\times$                                                                       & $\checkmark$                                                                     & $\checkmark$                                              \\
W-MSE        & $\checkmark$                                                                  & $\times$                                                                       & $\times$                                                                         & $\checkmark$                                              \\
CorInfoMax   & $\checkmark$                                                                  & $\checkmark$                                                                   & $\times$                                                                         & $\checkmark$                                              \\
I-VNE        & $\checkmark$                                                                  & $\checkmark$                                                                   & $\times$                                                                         & $\checkmark$                                              \\
MMCR         & $\checkmark$                                                                  & $\checkmark$                                                                   & $\times$                                                                       & $\checkmark$                                            \\
FroSSL (ours)       & $\checkmark$                                                                  & $\checkmark$                                                                   & $\checkmark$                                                                     & $\checkmark$                                              \\ \hline
\end{tabular}}
\end{table}

\section{The FroSSL Objective}
\label{sect:objective}


To motivate FroSSL, we begin by posing four desirable criteria of dimension-contrastive methods.

\begin{enumerate}
    \item \textbf{Invariant to Projection Rotations} We argue that dimension-contrastive methods should be invariant to rotations in the projections because the orientation of the covariance does not affect the relationships between principal components. In other words, redundancy in the embedding dimensions is invariant to the rotation of the embeddings. Thus the choices of $D_{\textrm{var}}$ and  $D_{\textrm{inv}}$ should be rotationally invariant as well.
    
    \item \textbf{Manipulates Eigenvalues Explicitly} Several works have shown that regularizing projection covariance eigenvalues in SSL can lead to reduced training time and improved downstream performance~\citep{halvagal2023implicit, kim2023vne, yerxa2024mmcr}. We provide empirical support for this in Section \ref{sect:efficient-learning-eigenvalue}.

    \item \textbf{Scales Quadratically in Batch Size and Dimension} The time complexity of the objective function scale \textit{at most} quadratically with respect to $N$ and $D$. This is often in opposition to the prior criteria which typically requires cubic eigendecomposition.

    \item \textbf{Scales Linearly in Views} The time complexity of the objective function should be linear in the number of views $V$. This is advantageous because recent work has shown that using more views can reduce training time and improve downstream performance~\citep{caron2020unsupervised, yerxa2024mmcr}. We provide empirical support for this in Sections \ref{sect:efficient-learning-multiview} and \ref{sect:experiments}. Any dimension-contrastive method with $D_{\textrm{inv}}$ that satisfies Equation $\eqref{eq:mean-target}$ fulfills this criterion.
\end{enumerate}

As shown in Table \ref{tab:dc-criteria-taxonomy}, no prior method meets all four criteria. We provide explanations in Appendix \ref{appendix:criteria-derivations}. To construct a method that fulfills all criteria, we modify the I-VNE objective function:
\begin{equation}
\label{eq:ivne}
\max \mathcal{L}_{\textrm{I-VNE}} = \sum_{v=1}^V \Tr{\Sigma_v \ln{\Sigma_v}} + \sum\limits_{v=1}^{V-1}\sum\limits_{r=v+1}^{V}  \frac{Z_v^T Z_r}{\schatten{Z_v}{2} \schatten{Z_r}{2}}
\end{equation}
The invariance term maximizes the cosine similarity between views. The variance term maximizes the von Neumann entropy of each view covariance matrix. The only criteria that I-VNE does not fulfill is being subcubic in batch size and dimension. This is due to the eigendecomposition needed to compute the matrix logarithm for the entropy. To begin addressing this, we first notice that the von Neumann entropy is a limit case of  matrix-based $\alpha$-order entropy~\citep{sanchezgiraldo2015measures, skean2023dime, hoyos2023representation}:
\begin{equation}
\label{eq:alphaentropy}
S_{\alpha}\left(\Sigma_v\right) = \frac{1}{1-\alpha}\log{\left[\Tr{\left( \Sigma_v ^{\alpha}\right) }\right]} = \frac{1}{1-\alpha} \log \left( \sum_i^{\min(N,D)} \lambda_i^\alpha (\Sigma_v)\right)
\end{equation}
Here, we do not require $\trace{\Sigma_{v}} =1$ as is typically required by $\alpha$-order entropy.  The von Neumann entropy is equivalent to $S_1(\Sigma_v)$ in the limit. Another special case is collision entropy, given by $S_2(\Sigma_v)$ below: 
\begin{equation}
\label{eq:frob-expanded}
S_{2}\left(\Sigma_v\right) = -\log\left(\sum_{i=1}^{\min(N, D)}{\lambda_i^2(\Sigma_v)}\right)  = -\log(\frobeniussq{\Sigma_v}) = -\log \sum_{i} \sum_j (\Sigma_v)_{ij}^2
\end{equation}
Notice how the left-hand side in the above equation explicitly uses the eigenvalues, while the right-hand side only uses matrix elements. This is made possible by the Frobenius norm, which offers an equivalency between a sum over eigenvalues and a sum over matrix elements. This has immediate impacts on the loss time complexity by relaxing the $O(\min(D,N)^3)$ eigendecomposition to the $O(\min(D,N)^2)$ Frobenius norm computation. The case of 2-order entropy is the only matrix-based $\alpha$-entropy which does not require eigendecomposition. One potential downside is $S_2(\Sigma_v)$ does not penalize outlying eigenvalues as heavily as in $S_1(\Sigma_v)$. This is akin to the difference between mean-absolute error and mean-squared error. However, this has no significant impact in our experiments.

The variance term $D_{\textrm{var}}$ for our objective will minimize the log Frobenius norm of normalized embeddings, causing the embeddings to spread out equally in all directions. Normalizing the embeddings is crucial because otherwise, minimizing the Frobenius norm will lead to trivial collapse. For the invariance term $D_{\textrm{inv}}$, we opt to use the mean-squared error between views. Our objective function FroSSL is given below:
\begin{equation}
\label{eq:frossl}
  \textrm{minimize   }  \mathcal{L}_{\textrm{FroSSL}} = 
 \sum\limits_{v=1}^{V} \log(\frobeniussq{\Sigma_v}) + \gamma \frobeniussq{Z_v - \overline{Z}}
\end{equation}
Note we simplify the pairwise mean-squared error via Equation \eqref{eq:mean-target}. Because the Frobenius norm is invariant to transposition, we can choose to compute either
$\frobeniussq{Z_v^T Z_v}$ or $\frobenius{Z_v Z_v^T}$ depending on if $D > N$. The former has time complexity $O(ND^2)$ while the latter has complexity $O(N^2 D)$. For consistency, we always use the former in our experiments. We provide
pseudocode in Appendix \ref{appendix:pseudocode}.



\subsection{The Role of the Logarithm}
\label{sect:logarithm}
The $\log$ in Equation \eqref{eq:frossl} ensures that the contributions of the variance term to the gradient of the objective function become self-regulated ($\frac{d\log{f(x)}}{dx} = \frac{1}{f(x)}\frac{d f(x)}{dx}$) with respect to the invariance term. We later compare the experimental performance of Equation \eqref{eq:frossl} with and without the logarithms, showing that using logarithms leads to a gain in performance. Prior work has shown that Equation \eqref{eq:frossl} with no logarithms causes dead neurons in the final encoder layer \cite{kim2023vne}.

\subsection{FroSSL is both Sample-contrastive and Dimension-contrastive}

It can be shown, up to an embedding normalization, that FroSSL is both dimension-contrastive and sample-contrastive. First, we provide formal definitions of dimension-contrastive and sample-contrastive SSL, following Garrido \etal~\cite{garrido2023duality}.

\begin{definition}[Dimension-Contrastive Method]
\label{def:dimension-contrastive}
An SSL method is said to be dimension-contrastive if it minimizes the non-contrastive criterion $\mathcal{L}_{nc}(Z) = \frobeniussq{Z^T Z - \textrm{diag}(Z^T Z)}$, where $Z \in \mathbb{R}^{N\mathrm{x}D}$ is a matrix of embeddings as defined above. This may be interpreted as penalizing the off-diagonal terms of the embedding covariance.
\end{definition}

\begin{definition}[Sample-Contrastive Method]
\label{def:sample-contrastive}
An SSL method is said to be sample-contrastive if it minimizes the contrastive criterion $\mathcal{L}_c(Z) = \\\frobeniussq{Z Z^T - \textrm{diag}(Z Z^T)}$. This may be interpreted as penalizing the similarity between pairs of different images.
\end{definition}

  Next, we use the duality of the Frobenius norm, given by $\frobenius{Z^T Z} = \frobenius{Z Z^T}$, to show that FroSSL satisfies the qualifying criteria of both dimension-contrastive and sample-contrastive methods. 

\begin{proposition}If every embedding dimension is normalized to have equal variance, then FroSSL is a dimension-contrastive method.  Proof in Appendix \ref{appendix:dimension-contrastive-proof}.
\label{prop:sear-dimension-contrastive} 
\end{proposition}

\begin{proposition}If every embedding is normalized to have equal norm, then FroSSL is a sample-contrastive method. Proof in Appendix \ref{appendix:sample-contrastive-proof}.
\label{prop:sear-sample-contrastive}
\end{proposition}

\begin{proposition}
    \label{prop:doubly}
    If the embedding matrices are doubly stochastic, then FroSSL is simultaneously dimension-contrastive and sample-contrastive.
\end{proposition}

Proposition \ref{prop:doubly} allows for interpreting FroSSL as either a sample-contrastive or dimension-contrastive method, up to a normalization of the data embeddings. The choice of normalization strategy is not important to the performance of an SSL method~\cite{garrido2023duality}. Unless otherwise specified, we only normalize the variance and not the embeddings. These same proof techniques can be used to show that TiCo, MMCR, I-VNE, and CorInfoMax also belong to both families~\citep{zhu2022tico, yerxa2024mmcr, kim2023vne, ozsoy2022self}. Additionally, variants of the dimension-contrastive VICReg have been proposed~\cite{garrido2023duality} that allow it to be rewritten as the sample-contrastive SimCLR. However, VICReg cannot be rewritten in such a way due to the hinge loss.

While Proposition \ref{prop:doubly} is interesting theoretically, it also offers empirical benefits to FroSSL. We examine overall wall-training time to reach competitive accuracies (\ref{sect:tradeoffs}), robustness to augmentations (\ref{sect:robustness}), and performance on little pretraining data (\ref{sect:low-data}). In all of these experiments, sample-contrastive methods outperform dimension-contrastive methods. FroSSL shares the advantages observed empirically in sample-contrastive methods.

\section{On Efficiency in Self-Supervised Learning}
\label{sect:training-dynamics}

It is well-known that traditional SSL algorithms need hundreds or thousands of epochs to reach competitive accuracies. To compare the efficiency of different SSL algorithms, we can borrow theoretical and practical tools from the broader field of algorithmic complexity. In the context of machine learning, there are two measurements of particular interest to practitioners: \textit{wall-time} needed to reach a given accuracy and VRAM \textit{space} used. The former can be decomposed into the atomic quantities of average \textit{wall-time per minibatch} and the \textit{number of epochs} needed to reach a given accuracy. To emphasize why this decomposition matters, consider a scenario where wall-time per minibatch differs between two methods but overall wall-time does not. In such a scenario, using the method with the slower minibatch wall-time is advantageous for using fewer disk reads and less distributed network traffic. This is not obvious without observing the atomic quantities. Note we are careful to specify ``epochs to reach a given accuracy'' rather than ``epochs to convergence''. One reason for this is that classical experiments in SSL train for a fixed number of epochs rather than until convergence. Another reason is algorithms that more quickly reach a target performance, such as FroSSL or I-VNE, do not necessarily converge in fewer epochs.

The design of an SSL algorithm is a balancing act between minibatch time, space, and number of epochs. While conversations involving minibatch time and space have been rarely discussed in the SSL literature, discussion about the number of epochs has seen renewed interest \citep{tong2023emp, simon2023stepwise, halvagal2023implicit}. However, if SSL algorithm design is indeed a balancing act of the three quantities above, then space and minibatch time deserve discussion too. Methods that boast reductions to one quantity may come with significant penalties to a different quantity. For example, dimension-contrastive methods use less space in practice than sample-contrastive methods, which prefer large minibatch sizes, or asymmetric methods like BYOL, which need an additional prediction network. However, the improved space efficiency comes at the cost of requiring a higher number of epochs~\citep{simon2023stepwise}. In Sections~\ref{sect:efficient-learning-eigenvalue} and \ref{sect:efficient-learning-multiview}, we discuss the advantages and drawbacks of two approaches to reducing the number of epochs. In Section \ref{sect:tradeoffs}, we compare a variety of SSL algorithms and visualize their time, space, and epoch tradeoffs.

\subsection{Reducing the Number of Epochs with Eigenvalue Dynamics}
\label{sect:efficient-learning-eigenvalue}

\begin{figure*}[!t]
\centering
\includegraphics[scale=0.35]{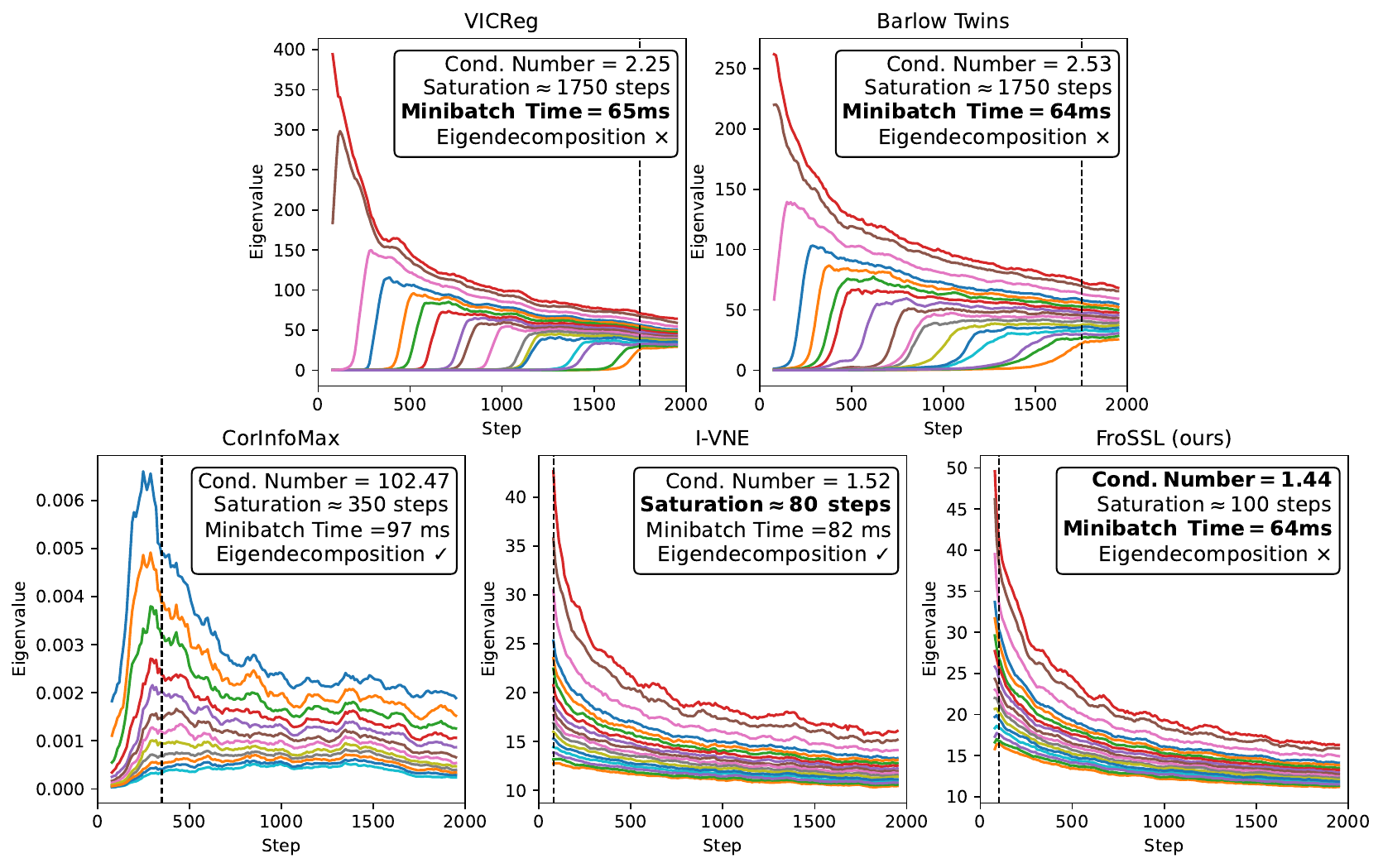}


\caption{\textbf{The choice of variance term, $D_{\textrm{var}}(\Sigma_{v}\Vert \mathbf{I})$,  has a significant impact on training dynamics}. Each subplot visualizes the trajectories of the top 20 eigenvalues of the embedding covariance matrix $\Sigma_{1}$ when trained with dimension-contrastive methods. These trajectories show how quickly $\Sigma_v$ converges to $\gamma I$, which has eigenvalues equal to $\frac{\gamma}{D}$.   VICReg, Barlow Twins, and CorInfoMax converge slowly. FroSSL and I-VNE have similar training dynamics, but FroSSL has significantly lower computational complexity because it avoids explicitly computing the eigendecomposition.}
\label{fig:stepwise-stl10}
\end{figure*}

Recent work has examined the training dynamics of SSL models~\citep{simon2023stepwise}. In particular, they find that the eigenvalues of the covariance exhibit \textit{stepwise} behavior, meaning that one eigendirection is learned at a time. This is readily seen in Figure \ref{fig:stepwise-stl10} for VICReg and Barlow Twins. This phenomenon contributes to slowness in SSL optimization with the smallest eigendirections taking the longest to be learned. Other work shows that high-rank representations lead to better downstream performances~\citep{garrido2023rankme}. It directly follows that if an SSL method requires a high number of epochs to learn high-rank representations, then it also needs a high number of epochs to learn useful representations. 

We hypothesize that by carefully choosing the variance term $D_{\textrm{var}}(\Sigma_{v}\Vert \mathbf{I})$ to reduce stepwise eigenvalue dynamics, useful representations can be learned more quickly. Indeed, this behavior has already been observed in several SSL objectives already. CorInfoMax optimizes the log-determinant of each covariance $\Sigma_v$, which is defined as the log of the product of the $\Sigma_v$ eigenvalues~\citep{ozsoy2022self}. IsoLoss uses $\Sigma_v$ eigenvalues as learning rate multipliers to equalize the convergence rate of different eigenmodes~\citep{halvagal2023implicit}. MMCR optimizes the nuclear norm of the average view embedding, which is defined as the sum of the singular value magnitudes~\citep{yerxa2024mmcr}. I-VNE optimizes the von Neumann entropy of $\Sigma_v$, which is equal to the Shannon entropy of the $\Sigma_v$ eigenvalues~\citep{kim2023vne}. 

It is straightforward to show that FroSSL also directly influences the covariance eigenvalue dynamics. However, FroSSL is unique from prior methods because it does so while avoiding explicit eigendecomposition. This can be seen from Equation \ref{eq:frob-expanded}. Additionally, using the Frobenius norm eliminates numerical instabilities typically associated with eigendecomposition~\citep{dang2018eigendecomposition}.

 To highlight the existence and remedy of stepwise phenomena in practical scenarios, we create an experiment similar to the one used by Simon \etal~\cite{simon2023stepwise}. In Figure \ref{fig:stepwise-stl10}, we plot the trajectories of the top 20 eigenvalues of $\Sigma_{1}$ when trained with different dimension-contrastive objectives. For all SSL objectives, a ResNet-18 was trained for 5 epochs on STL-10 using SGD with $lr=0.01$ and a batch size of $256$.  Further details are given in Appendix \ref{sect:stepwise-details}. 
 
 The eigenvalues trajectories show how quickly $\Sigma_v$ is approaching $\gamma I$, which has eigenvalues equal to $\frac{\gamma}{D}$. We say that an objective is saturated once the stepwise learning phase is ended. This is marked as the step where $\lambda_{20}$ has increased from zero and started decreasing. It is clear to see that SSL objectives that directly influence eigenvalues, namely CorInfoMax, I-VNE, and FroSSL, saturate much quicker than the others. Interestingly, the condition number for CorInfoMax, computed as $\frac{\lambda_1}{\lambda_{20}}$, is much larger than any other tested method. We hypothesize this is due to the choice of the $\epsilon$ hyperparameter for the regularization term when computing the determinant as $\det (\Sigma_1 + \epsilon I)$.

\subsection{Reducing the Number of Epochs by Using More Views}
\label{sect:efficient-learning-multiview}

\subsubsection{Multiview with 3 or More Views}
In contrastive learning, using more views has been shown to have significant impacts on representation quality and downstream performance~\cite{tian2020contrastive}. In SSL, using more augmentations for each image has the effect of averaging out noise from the mean embedding, which acts as a target for many SSL objectives as shown in Equation \eqref{eq:mean-target}. This differs from increasing the batch size, which would instead average out noise across samples and not across targets. While using more views is promising, it has not seen widespread adoption in self-supervised learning. This is in part due to many sample-contrastive methods being quadratic in the number of views. However, this problem is circumvented for the dimension-contrastive methods shown in Table \ref{tab:dc-taxonomy}, which are instead linear in the number of views. One such method, W-MSE, has shown performance improvements when the number of views is increased from 2 to 4~\cite{ermolov2021whitening}. Interestingly, MMCR is constant in the number of views because it operates only on the mean embedding~\cite{yerxa2024mmcr}.

\begin{table}[!t]
\centering
\caption{Comparison of the time/space/epoch tradeoffs for SSL algorithms trained on STL-10. FroSSL with 8 views achieves 80\% top-1 accuracy in the least wall-time.}
\label{tab:time-complexities}
\scalebox{0.7}{
\begin{tabular}{r|ccc|ccc|ccc|ccc}
\hline
                              & SimCLR         & MoCo v2   & BYOL  & VICReg       & Barlow       & CorInfoMax & \multicolumn{3}{c|}{MMCR}              & \multicolumn{3}{c}{FroSSL (ours)}       \\ \hline
Num. Views                    & 2              & 2        & 2     & 2            & 2            & 2          & 2            & 4          & 8          & 2        & 4        & 8                 \\
Loss Time Complexity          & \multicolumn{3}{c|}{$O(V^2 N^2)$} & $O(V D^2)$   & $O(V^2 D^2)$ & $O(V D^3)$ & \multicolumn{3}{c|}{$O(\min(D, N)^3)$} & \multicolumn{3}{c}{$O(V \min(D, N)^2)$} \\ \hline
VRAM Space (GB)               & \textbf{1.6}   & 2.8      & 2.0   & \textbf{1.6} & \textbf{1.6} & 1.7        & 1.7          & 2.9        & 5.3        & 1.7      & 2.9      & 5.3               \\
Minibatch Wall-time (ms)      & \textbf{60}    & 79       & 76    & 65           & 64           & 97         & 71           & 108        & 187        & 64       & 105      & 187               \\
Number of Epochs to 80\% Acc  & 347            & 180      & 187   & 360          & 370          & 405        & 380          & 211        & 63         & 290      & 144      & \textbf{55}       \\
Wall-time to 80\% Acc (hours) & 2.4            & 1.6      & 1.6   & 2.7          & 2.7          & 4.5        & 3.1          & 2.6        & 1.3        & 2.1      & 1.7      & \textbf{1.2}      \\ \hline
\end{tabular}}
\end{table}

\begin{table}[!ht]
    \caption{
    Top-1 accuracies on STL-10 using an online linear classifier during training for specific numbers of epochs (left) and specific elapsed times (right). 
    }
    
\label{tab:stl10_per_epoch}
    \begin{subtable}{.45\linewidth}
    \centering
\scalebox{0.65}{
\begin{tabular}{lccccc}
\hline
                                  & \multicolumn{5}{c}{Epochs}        \\
\multicolumn{1}{c}{Method}        & 3    & 10   & 30   & 50 & 100 \\ \hline

\quad SimCLR       & 40.7 & 44.8 & 61.5 & 66.2 & 70.1 \\
\quad MoCo v2         & 24.6 & 45.0 &  63.8    &  69.4    & 75.2     \\
\quad BYOL         & 28.8 & 32.7 & 59.6 & 64.7 & 70.6 \\
\quad VICReg       & 43.6 & 51.1 & 61.2 & 67.5 &  71.1    \\
\quad Barlow Twins & 32.1 & 46.6 & 62.0 & 62.6 & 69.0 \\
\quad CorInfoMax     & 39.0 & 49.1 & 58.0 & 62.5 & 66.2 \\ \hline \\[-1em]

\quad MMCR (2 views)        & 39.6 & 53.3  & 62.8 & 63.3 & 67.0 \\

\quad MMCR (4 views)        & 46.0 & 61.5 & 70.2 & 71.5 & 75.7 \\

\quad MMCR (8 views)        & \textbf{51.1} & 64.7 & 72.9 & 77.2 & 79.4 \\

\hline \\[-1em]

\quad FroSSL (2 Views)         & 44.8 & 56.9 & 64.8 & 67.1 & 72.0 \\

\quad FroSSL (4 Views)         & 49.3 & 60.7 & 70.3 & 67.1 & 76.9 \\

\quad FroSSL (8 Views)         & 47.6 &\textbf{ 65.5} & \textbf{74.5} & \textbf{78.4} & \textbf{81.8} \\ \hline

\end{tabular}
}

    \end{subtable}%
    \hfill
\begin{subtable}{.45\linewidth}
\scalebox{0.65}{
\begin{tabular}{lccc}
\hline
                                  & \multicolumn{3}{c}{Training  Wall-Time (min.)}        \\ 
\multicolumn{1}{c}{Method}        & 10    & 30   & 60  \\ \hline

\quad SimCLR    &   61.5 & 68.8 & 73.9 \\
\quad  MoCo v2    &  57.4 & 70.0 & 76.2 \\
\quad BYOL     & 50.2 & 65.3 & 75.0 \\
\quad VICReg   &   63.4 & 70.3 & 72.9 \\
\quad  Barlow Twins  &  55.5 & 66.1 & 68.0 \\
\quad CorInfoMax  &   56.1 & 64.9 & 65.6 \\

\hline \\[-1em]

\quad MMCR (2 views)        & 54.7  & 68.4 &  70.9\\

\quad MMCR (4 views)        & \textbf{69.8}   & \textbf{74.7}  &  76.6 \\

\quad MMCR (8 views)        & 64.7   & 73.5   & 78.0 \\

\hline \\[-1em]

\quad FroSSL (2 Views)         &  63.4  &    69.7 &    74.9 \\

\quad FroSSL (4 Views)         & \textbf{68.5} &  73.7  &   76.3 \\

\quad FroSSL (8 Views)    &    58.6   &  74.1   &  \textbf{79.0} \\ \hline

\end{tabular}
}
    \end{subtable} 
\end{table}

\subsubsection{Multi-Patch and Multi-Crop Methods}
An approach in a different vein is to extract and augment image patches to serve as views, rather than using full images. EMP-SSL has shown that this drastically reduces the number of epochs and overall wall-time needed to reach competitive accuracies by utilizing a bag-of-features model that embeds hundreds of small augmented patches per image \cite{tong2023emp, chen2022bag}. However, EMP-SSL comes at the cost of major penalties to time-per-minibatch and space in both training time and inference time.

As an alternative to using full-sized images or tiny patches for each view, multi-crop methods strike a balance~\cite{caron2020unsupervised}. A certain number of views are full-sized images while the remaining views are smaller crops. A typical setup for ImageNet is using two $224 \times 224$ views and six $96 \times 96$ views. These approaches differ from our experiments which use all full-sized views with FroSSL. However, we expect that FroSSL should work well as an objective function for these paradigms too.

\subsection{Exploring Time, Space, and Epoch Tradeoffs}
\label{sect:tradeoffs}

We now compare the efficiency of different SSL algorithms. We train a ResNet-18 for 500 epochs on STL-10 and measure the number of epochs needed to reach a top-1 accuracy of $80\%$. This threshold of $80\%$ was chosen because all methods achieve that accuracy within 500 epochs. We used $N=256$ and $D=1024$ for all methods. Because these models were trained on a distributed cluster, it is important to compensate for different compute when measuring minibatch wall-time. In particular, we measure minibatch time by averaging over 2000 iterations of training on one NVIDIA A5000 GPU. We measure VRAM space as the maximum space requested by the training script. We calculate wall-time to $80\%$ accuracy by multiplying minibatch time, epochs, and iterations per epoch. 

In Table \ref{tab:time-complexities} we show the resources needed for each SSL objective. There are several observations to glean from this table. First, increasing the number of views reduces epochs and overall wall-time, even though space and minibatch time become larger. FroSSL with 8 views reaches 80\% top-1 accuracy faster than any other tested method. Second, asymmetric methods require the least overall wall-time for any method using 2 views, at the cost of space. We hypothesize this is due to enhanced training stability from momentum encoders. Third, doubling the number of views does not necessarily double the minibatch wall-time. This is because some parts of the training script, such as data loading and logging, do not get slower as the number of views increases. In Table \ref{tab:stl10_per_epoch}, we show top-1 accuracies over epochs and over time. In both scenarios, FroSSL with 8 views has the highest accuracy after training is finished. 

The impact of loss time complexity on minibatch time is more apparent at extreme batch sizes and dimensionalities. We show the VRAM space and minibatch time for $N=1024$ and $D=1024$ in Table \ref{tab:times-bigbatchsize}. In this setting, FroSSL has 33\% and 14\% faster minibatch times than MMCR when both have 2 or 8 views, respectively. Additionally, when compared to MoCo v2 and BYOL, FroSSL with 2 views has a 20\%  faster minibatch time and uses 20\% less space. 

\section{Experimental Results}
\label{sect:experiments}
\begin{table*}[!t]
\centering
\caption{Comparison of SSL methods on small datasets. All CorInfoMax and MMCR results are from our implementation. All Tiny ImageNet and STL-10 results are from our implementation. CIFAR-10 and CIFAR-100 results are reported from~\cite{turrisi2022solo, ermolov2021whitening}. IN-100 baseline results are from \cite{turrisi2022solo}. We observed negligible improvements from using more views for FroSSL on the CIFAR datasets; we used the 2-view CIFAR accuracies to compute 4/8 view averages. An asterisk (*) denotes Tiny ImageNet results where weak augmentations outperformed strong ones. Results within 
0.5\% of best are \textbf{bolded}.}
\label{tab:small_datasets}
\scalebox{0.7}{
\begin{tabular}{lcccccc}
\hline
\multicolumn{1}{c}{Method}     & CIFAR-10             & CIFAR-100            & STL-10 & Tiny-IN & IN-100 & Average              \\ \hline
\textbf{Sample-Contrastive}    &                      &                      &         &  & &           \\
\quad SimCLR                         & 91.8                 & 65.8                 &     85.9     & 41.9 & 77.6 &  72.6     \\
\quad SwAV                           & 89.2                 & 64.9                 & 82.6  & 41.2 & 74.3 &   70.5               \\
\quad MoCo v2                        & \textbf{92.9}        & 69.9                 & 83.2 &41.9& 79.3 & 73.4 \\
\textbf{Asymmetric Network}    & & &  & & &\\
\quad SimSiam                        & 90.5                 & 66.0                 &   88.5  & \textbf{45.6*} & 78.7 & 73.9               \\
\quad BYOL                           & \textbf{92.6}                 & \textbf{70.5}                 &   88.7  & 40.1 &  \textbf{80.3}  & 74.4 \\
\quad DINO                           & 89.5                 & 66.8                 & 78.9 &34.9 & 78.9 & 69.8 \\
\textbf{Dimension-Contrastive} &  &  &  & & &\\

\quad VICReg                         & 92.1                 & 68.5                 &  85.9  & 37.5 & 79.4 & 65.8               \\
\quad Barlow Twins                   & 92.1                 & \textbf{70.9}        &    85.0          &  \textbf{45.3} &\textbf{ 80.2} & 74.7  \\
\quad W-MSE 2                        & 91.6                 & 66.1                 &   72.4    &28.8* & 69.0 & 65.6               \\
\quad CorInfoMax                        &    \textbf{92.6}            &         69.7         &  83.1  & 43.9 & 74.7 & 72.8              \\
\quad I-VNE & 89.7 & 65.7 & 87.4 & \textbf{45.2} & 77.6 & 73.1\\
\hline \\[-1em]

\quad MMCR (2 views)                     &   88.6            &   65.8             &   84.3   & 41.2 & 76.7 &  71.3            \\

\quad MMCR (4 views)                         &  89.6           &   67.3              &  88.2   & 42.8 & 78.8 &    73.3          \\

\quad MMCR (8 views)                         &  89.3              &      68.3          &   90.3   & 43.2 & \textbf{80.3} &   74.2           \\

\hline \\[-1em]

\quad FroSSL (2 views) [no log]                           &         88.9    &    62.3       &   82.4           & 36.4 & 78.3 & 69.7 \\

\quad FroSSL (2 views)                         &  \textbf{92.8}           &  \textbf{70.6}            &   87.3          & 44.2 & 78.2 & 74.6 \\

\quad FroSSL (4 views)                          &   -         &   -          &    90.0       & \textbf{45.3} & 79.4 & 75.6 \\

\quad FroSSL (8 views)                          &  -       &          -   &  \textbf{90.9}          & \textbf{45.3} & \textbf{79.8} & \textbf{75.9} \\

\hline
\end{tabular}}
\end{table*}


\label{sect:small-datasets}
In this section, we train a ResNet-18 on CIFAR-10~\citep{krizhevsky2009learning}, CIFAR-100, STL-10~\citep{coates2011analysis}, Tiny ImageNet~\citep{le2015tiny}, and ImageNet-100~\citep{tian2020contrastive}. Our implementation is based on the solo-learn SSL framework~\citep{turrisi2022solo}. 

In Table \ref{tab:small_datasets}, we show linear probe evaluation results on these datasets. It is readily seen that FroSSL learns competitive representations in comparison to other SSL methods. The implementation details can be summarized as:

\begin{itemize}
\item \textbf{Optimizer}  The backbone uses LARS optimizer~\citep{you2017large} with an initial learning rate of $0.3$,  weight decay of 1e-6, and a warmup cosine learning rate scheduler. The linear probe uses the SGD optimizer~\citep{kingma2014adam} with an initial learning rate of $0.3$, no weight decay, and a step learning rate scheduler with decreases at 60 and 80 epochs.

\item \textbf{Epochs} 
For CIFAR-10 and CIFAR-100, we pretrain the backbone for 1000 epochs. For STL-10, we pretrain for 500 epochs. For Tiny ImageNet, we pretrain for 800 epochs. For ImageNet-100, we pretrain for 800 epochs. All linear probes were trained for 100 epochs.

\item  \textbf{Hyperparameters}
A batch size of N=256 is used for all datasets, except for Tiny ImageNet which used N=512. For FroSSL, we used $\gamma=1.4$ for 2 views and $\gamma=2$ for 4 and 8 views. We used an MLP with output dimension $D=1024$ for FroSSL.  Details about augmentations and method-specific hyperparameters are given in Appendix \ref{sect:linear-experimental-details}.
\end{itemize}

\subsection{Robustness to Augmentations}
\label{sect:robustness}
We trained ResNet-18 models on Tiny ImageNet using both weak and strong augmentations. Weak augmentations had Gaussian blur probabilities (0.5, 0.5) and solarization probabilities (0, 0) for each view. Strong augmentations had Gaussian blur probabilities (1.0, 0.1) and solarization probabilities (0.2, 0). As shown in Table \ref{tab:tiny-aug}, FroSSL is more robust to changes in augmentations than any other dimension-contrastive method.

\subsection{Performance In Low Data Regime}
\label{sect:low-data}
We trained ResNet-18 models on ImageNet-1K~\citep{deng2009imagenet} for 100 epochs using only 10\% of the data and evaluated them using the standard linear probe. Note that limited data was used in both pretraining and evaluation. As shown in Table \ref{tab:low-data}, FroSSL achieves a better downstream performance on limited data than any other tested method. 






\begin{table}[!t]
    \begin{minipage}{.45\linewidth}
\caption{The top-1\% accuracies after training on Tiny-ImageNet using weak or strong augmentations.}
\label{tab:tiny-aug}
\setlength{\tabcolsep}{8pt}
\scalebox{0.7}{
\begin{tabular}{lccc}

\hline
Method       & Weak                                                                                                    & Strong      & $\Delta \%$                                                                                                        \\ \hline

SimCLR       & 39.5                                                                                                                      & 41.9 & 2.4       \\
SwAV       &       39.9                                                                                                                &   41.2  & 1.5 \\
MoCo v2       &       40.9                                                                                                                &   41.9  & 1.0 \\
\hline
SimSiam       &       45.6                                                                                                                &  39.7   & -5.9 \\

BYOL       &       39.4                                                                                                                &   40.1  & 0.7 \\

DINO       &       32.2                                                                                                               &   34.9 & 2.7  \\

\hline
VICReg       &       18.1                                                                                                                &   37.5  & 19.6 \\

Barlow Twins & 36.8                                                                                                                      & 45.3  & 8.5                                                                                                                      \\
CorInfoMax       &       33.1                                                                                                                &   43.9 & 10.8   \\
\hline
MMCR (2 views)       &       24.2                                                                                                                &   41.2 & 17.0  \\

FroSSL (2 views)       & 39.4                                                                                                                      & 44.2 & 4.8 \\ \hline       
\end{tabular}}
    \end{minipage}%
    \hfill
    \begin{minipage}{.45\linewidth}
    \centering
\caption{Accuracies after pretraining on ImageNet-1k for 100 epochs with only 10\% of data.}
\label{tab:low-data}
\begin{tabular}{lcc}
\hline
                 & Top-1 & Top-5 \\ \hline
SimCLR           & 31.1  & 56.6  \\
BYOL             & 12.7  & 29.1  \\
SimSiam          & 22.7  & 46.3  \\
Barlow Twins     & 23.6  & 46.9  \\\hline
FroSSL (2 Views) & 33.4  & 59.1  \\
FroSSL (8 Views) & 38.2  & 64.1  \\ \hline
\end{tabular}
    \end{minipage} 
\end{table}

\section{Conclusion}
We introduced FroSSL, a self-supervised learning method that can be seen as both sample- and dimension-contrastive. We showed that FroSSL enjoys the simplicity of dimension-contrastive methods while achieving the empirical advantages of sample-contrastive methods. In particular, we discovered that FroSSL can achieve substantially stronger performance than alternative SSL methods when trained with less overall wall-time. To better understand why this is happening, we presented empirical results based on eigenvalue trajectories. We also demonstrated the effectiveness of FroSSL through extensive experiments on standard datasets.


\section*{Acknowledgements}

This research is based upon work supported in part by the Office of the Director of National Intelligence (ODNI), Intelligence Advanced Research Projects Activity (IARPA), via  2021-2011000005 and the Office of the Under Secretary of Defense for Research and Engineering under award number FA9550-21-1-0227. The views and conclusions contained herein are those of the authors and should not be interpreted as necessarily representing the official policies, either expressed or implied, of ODNI, IARPA, the U.S. Department of Defense, or the U.S. Government. The U.S. Government is authorized to reproduce and distribute reprints for governmental purposes notwithstanding any copyright annotation therein.

\bibliographystyle{unsrtnat}
\bibliography{main}

\appendix
\newpage

\section{Pseudocode}
\label{appendix:pseudocode}
\subsection{Pseudocode for FroSSL - 2 Views}
\lstset{
    basicstyle=\ttfamily\footnotesize,
    breaklines=true,
    morecomment=[l][\color{red}]{\#}
}
\begin{lstlisting}[frame=single, linewidth=13cm]
for x in loader:
    # augment the image
    x_a, x_b = augment(x)

    # pass through network f to get embeddings
    z_a = f(x_a)
    z_b = f(x_b)
    N, D = Z_a.shape

    # center embeddings
    Z_a = Z_a - Z_a.mean(0)
    Z_b = Z_b - Z_b.mean(0)

    # normalize dimensions to sqrt(D) std.
    Z_a = (D**0.5) * (Z_a / Z_a.norm())
    Z_b = (D**0.5) * (Z_b / Z_b.norm())

    # calculate invariance (MSE) term
    invariance_loss = MSELoss(Z_a, Z_b)

    # calculate variance (Frobenius norm) term
    frobenius_a = torch.log(torch.norm(Z_a.T @ Z_a, ord='fro'))
    frobenius_b = torch.log(torch.norm(Z_b.T @ Z_b, ord='fro'))
    variance_loss = frobenius_a + frobenius_b 

    # FroSSL loss
    loss = gamma*invariance_loss + variance_loss
    loss.backward()
    optimizer.step()
\end{lstlisting}
\newpage
\subsection{Pseudocode for FroSSL - V Views}
\label{sect:pseudocodemulti}
\lstset{
    basicstyle=\ttfamily\footnotesize,
    breaklines=true,
    morecomment=[l][\color{red}]{\#}
}
\begin{lstlisting}[frame=single, linewidth=12.5cm]
for x in loader:
    # augment the image
    X_list = [augment(x) for v in range(V)]

    # pass through network f to get embeddings
    Z_list = [f(x) for x in X_list]
    N, D = Z_list[0].shape

    # normalize embeddings
    Z_list = [F.normalize(z) for z in Z_list]
    Z_mean = torch.mean(torch.stack(Z_list), dim=0)

    total_loss = 0
    for Z_v in Z_list:
        # calculate invariance (MSE) term
        invariance_loss = MSELoss(Z_v, Z_mean)
    
        # calculate variance (Frobenius norm) term quickly
        if N > D:
            cov = view_embeddings.T @ view_embeddings
        else:
            cov = view_embeddings @ view_embeddings.T
        cov = cov / torch.trace(cov)
        fro_norm = torch.linalg.norm(cov, ord='fro')
        
        # bring frobenius square outside log
        variance_loss = -2*torch.log(fro_norm)
        

        loss = gamma*invariance_loss + variance_loss
        total_loss += loss

    total_loss.backward()
    optimizer.step()
\end{lstlisting}
\newpage

\section{Derivations for Table 1}
\label{appendix:derivations}
Here we derive the $D_{var}$ and $D_{inv}$ terms for each row in Table 1. We would like to note that while it is tempting to include works like Barlow Twins and MMCR in the framework of Table \ref{tab:dc-taxonomy}, they do not fit neatly as they are calculated between views rather than on individual views. A concurrent work, Matrix-SSL~\cite{zhangmatrixssl}, introduces a loss function that could also be represented in the framework of Table \ref{tab:dc-taxonomy}. However, we leave that analysis to future work.

\subsection{FroSSL}
In FroSSL, the minimization of the Frobenius norm brings the embedding covariance eigenvalues towards $\frac{c}{D}$, where c is some scaling factor. These eigenvalues are the same as the eigenvalues of $cI$. Thus the goal of the variance term in FroSSL is to bring each view covariance towards a unitary transformation of $I$. In other words, we are minimizing some distance between $I$ and the covariance. This distance is, in fact, the Renyi relative entropy of order 2. Below we start with the 2-order Petz-Renyi relative entropy and derive the FroSSL variance term.

$$S_2(A||B) = \ln \left( \trace{A^2 B} \right)$$
$$S_2(\Sigma_v||I) = \ln \trace{\Sigma_v^2} = \ln \frobeniussq{\Sigma_v} $$

\subsection{I-VNE}
The intuition for I-VNE follows similarly to FroSSL. The only difference is the choice of relative entropy distance between $\Sigma_v$ and $I$. Here $S_1$, or von Neumann relative entropy, instead of $S_2$ as in FroSSL.

\subsection{VICReg} 
The distance between the covariance and identity is explicitly defined in the VICReg objective function. We simply use its variance terms and invariance terms for $D_{\textrm{var}}$ and $D_{\textrm{inv}}$, respectively.

\subsection{CorInfoMax}
First, notice that if we use variance-normalized features, we have $\trace{\Sigma_v + \epsilon \mathbf{I}} = D$. If we instead use sample-normalized features, then we have $\trace{\Sigma_v + \epsilon \mathbf{I}} = N$. In either case, the trace term remains constant and thus is not relevant for the optimization. The only non-constant term that changes is $\log\det(\Sigma_v + \epsilon \mathbf{I})$ which directly corresponds to the CorInfoMax loss.

\subsection{W-MSE}
Because the projections are explicitly whitened, we automatically get $D_{var}=0$. The choice of $D_{inv}$ comes straightforwardly from the loss.

\section{Desirable Dimension-Contrastive Criteria}
\label{appendix:criteria-derivations}
In this section, we go over each dimension-contrastive method and discuss why it meets or fails to meet the four desirable criteria laid out in Section 3.

\subsection{Barlow Twins}
\subsubsection{Invariant to Projection Rotation $\times$} The objective function explicitly tries to make the embedding cross-covariance identity. Rotations of identity are not suitable, even though they offer the same low redundancies as identity.  Thus this criteria is not met.

\subsubsection{Manipulates Eigenvalues Explicitly $\times$}
The objective function is almost, but not quite, a Frobenius norm. It is not clear how the objective function could be written in terms of eigenvalues.

\subsubsection{Scales Quadratically in Batch Size and Dimension $\checkmark$} The objective function is a sum of matrix elements, which scales in $O(D^2)$.

\subsubsection{Scales Linearly in Views $\times$} Because Barlow Twins is based on the cross-covariance, it must compute all pairwise cross-covariances as the number of views increases. Thus it scales quadratically in views.

\subsection{VICReg}
\subsubsection{Invariant to Projection Rotation $\times$} 
Same reasoning as with Barlow Twins, but with covariance matrices instead of cross-correlations.

\subsubsection{Manipulates Eigenvalues Explicitly $\times$}
Same reasoning as Barlow Twins.

\subsubsection{Scales Quadratically in Batch Size and Dimension $\checkmark$} Same reasoning as Barlow Twins.

\subsubsection{Scales Linearly in Views $\checkmark$} 
Uses the covariance for each view independently, rather than computing all pairwise cross-correlation matrices as in Barlow Twins. The mean-square error can be reduced to linear by Equation 3 in the main text.

\subsection{W-MSE}
\subsubsection{Invariant to Projection Rotation $\checkmark$} 
The whitening operation achieves this.

\subsubsection{Manipulates Eigenvalues Explicitly $\times$}
The objective function only tries to minimize the distance between the views.

\subsubsection{Scales Quadratically in Batch Size and Dimension $\times$} The whitening operation is computed in $O(D^3)$.

\subsubsection{Scales Linearly in Views $\checkmark$} The whitening operation is done for each view, and the mean-square error can be reduced to linear by Equation 3 in the main text.

\subsection{CorInfoMax}
\subsubsection{Invariant to Projection Rotation $\checkmark$} 
Determinant is invariant to rotation.

\subsubsection{Manipulates Eigenvalues Explicitly $\checkmark$}

The objective function uses the log of the covariance determinant, which is defined as the product of eigenvalues.

\subsubsection{Scales Quadratically in Batch Size and Dimension $\times$} Determinant can only be computed in $O(D^3)$.

\subsubsection{Scales Linearly in Views $\checkmark$} Computes a determinant for each view. The mean-square error can be reduced to linear by Equation 3 in the main text.

\subsection{I-VNE}
\subsubsection{Invariant to Projection Rotation $\checkmark$} The von Neumann entropy is invariant to rotations.

\subsubsection{Manipulates Eigenvalues Explicitly $\checkmark$}
The von Neumann entropy of a matrix is defined as the Shannon entropy of its eigenvalues.

\subsubsection{Scales Quadratically in Batch Size and Dimension $\times$} 
The von Neumann entropy requires explicit eigendecomposition, which is $O(\min(N,D)^3)$.

\subsubsection{Scales Linearly in Views $\checkmark$}
Computes an entropy for each view. The cosine similarity can be reduced to linear by Equation 3 in the main text.

\subsection{MMCR}
\subsubsection{Invariant to Projection Rotation $\checkmark$} The nuclear norm is invariant to rotations.

\subsubsection{Manipulates Eigenvalues Explicitly $\checkmark$}
The nuclear norm is defined as the sum of the magnitudes of singular values.

\subsubsection{Scales Quadratically in Batch Size and Dimension $\times$} 
The nuclear norm requires explicit eigendecomposition, which is $O(\min(N,D)^3)$.

\subsubsection{Scales Linearly in Views $\checkmark$}
Interestingly, MMCR is constant in the number of views. The objective function is computed on the average view embedding, rather than each view independently. 

\subsection{FroSSL}
\subsubsection{Invariant to Projection Rotation $\checkmark$} The Frobenius norm is invariant to rotations.

\subsubsection{Manipulates Eigenvalues Explicitly $\checkmark$}
The Frobenius norm can be defined as the sum of the squared eigenvalues.

\subsubsection{Scales Quadratically in Batch Size and Dimension $\checkmark$} 
The Frobenius norm can be computed as a sum over matrix elements in $O(\min(D^2, N^2))$.

\subsubsection{Scales Linearly in Views $\checkmark$}
A Frobenius norm is computed for each view. The mean-square error can be reduced to linear by Equation 3 in the main text.

\section{Experimental Details}
\subsection{Stepwise Convergence Experimental  Details}
\label{sect:stepwise-details}
We trained ResNet-18 on STL10 using SGD with $lr=0.01$ and a batch size of $256$. Training occurred for only $5$ epochs ($\approx$2000 steps) because we were interested in stepwise behaviors early during training. For all methods except CorInfoMax, a projector dimensionality of $D=1024$ was used.

\begin{itemize}
\item \textbf{Barlow Twins} {We used $\lambda=0.05$ as recommended by \cite{zbontar2021barlow}.}
\item \textbf{VICReg} {We used $\lambda=25$, $\mu=25$, $\nu=1$ as recommended by \cite{bardes2022vicreg}.}
\item \textbf{CorInfoMax} {We used temperature $\alpha=500$, $\epsilon=1e-6$, $\mu=1e-2$, $D=256$ following the STL-10 recommendations of \cite{ozsoy2022self}. }
\item \textbf{I-VNE} {The original work does not mention choices of tradeoffs between the invariance and variance terms \cite{kim2023vne}. Therefore, we equally weight the invariance and variance terms. We find this works well in practice}
\item \textbf{FroSSL} {We used an invariance weight $\gamma=1.4$.}
\end{itemize}

\subsection{Linear Probe Experimental Details}
\label{sect:linear-experimental-details}
\subsubsection{Hyperparameters}
We follow the guidance of \cite{turrisi2022solo}, \cite{ozsoy2022self}, and \cite{yerxa2024mmcr} for selecting baseline hyperparameters. Our code contains .yaml files with the exact hyperparameter configurations we used on each dataset.

\subsubsection{Augmentations}
For datasets other than Tiny ImageNet, we use the default symmetric augmentations provided in \cite{turrisi2022solo} for FroSSL. These:

\begin{itemize}
    \item Random Resized Crop - Crop scale ranges from 0.08 to 1.0, then resized to 32x32 for CIFAR, 96x96 for STL-10, 224x224 for ImageNet-100
    \item Color Jitter with probability 0.8 and (brightness, contrast, saturation, hue) values of (0.8, 0.8, 0.8, 0.2)
    \item Grayscale with probability 0.2
    \item Gaussian blur with probability 0.5
    \item Horizontal flip with probability 0.5
\end{itemize}

For Tiny ImageNet, we use the asymmetric augmentations provided in \cite{ozsoy2022self}. These are shown below. Because these were originally designed for methods using 2 views, we used equal numbers of each view type for multiview methods.

\begin{itemize}
    \item Random Resized Crop - Crop scale ranges from 0.08 to 1.0, then resized to 64x64
    \item Color Jitter with probability 0.8 and (brightness, contrast, saturation, hue) values of (0.4, 0.4, 0.2, 0.1)
    \item Grayscale with probability 0.2
    \item Gaussian blur with probability 1.0 in view 1 and probability 0.1 in view 2
    \item Solarization with probability 0 in view 1 and probability 0.2 in view 2
    \item Horizontal flip with probability 0.5
\end{itemize}

\section{Proofs}
\subsection{Proof of Proposition 1}
\label{appendix:dimension-contrastive-proof}
WLOG, we prove the following for 2 views. We start with rewriting the $\argmin$ of Equation 7 as such:
    $$\argmin_{Z_1, Z_2} \mathcal{L}_{\textrm{FroSSL}} =  \argmin_{Z_1, Z_2} \left[ \log(\frobeniussq{Z_1^T Z_1}) + \log(\frobeniussq{Z_2^T Z_2}) + \mathcal{L}_{\textrm{MSE}}(Z_1, Z_2) \right]$$
    $$\quad \quad \Hquad= \argmin_{Z_1, Z_2} \left[ \frobeniussq{Z_1^T Z_1} + \frobeniussq{Z_2^T Z_2} + \mathcal{L}_{\textrm{MSE}}(Z_1, Z_2) \right]$$
        Assume the normalization step normalizes each embedding dimension to have unit variance. Then both covariance matrices have $1$ in each diagonal element.
        $$= \argmin_{Z_1, Z_2} \left[ \frobeniussq{Z_1^T Z_1 - \textrm{diag}(Z_1^T Z_1)} +  \frobeniussq{Z_2^T Z_2 - \textrm{diag}(Z_2^T Z_2)} + 2D + \mathcal{L}_{\textrm{MSE}}(Z_1, Z_2) \right]$$
        $$=  \argmin_{Z_1, Z_2} \left[ \mathcal{L}_{nc}(Z_1) + \mathcal{L}_{nc}(Z_2) + 2D + \mathcal{L}_{\textrm{MSE}}(Z_1, Z_2) \right]$$
Thus we have that the embeddings that minimize FroSSL also minimize the non-contrastive losses $\mathcal{L}_{nc}$ for both views.

\subsection{Proof of Proposition 2}
\label{appendix:sample-contrastive-proof}
   WLOG, we prove the following for 2 views. With the duality of the Frobenius norm, we rewrite Equation 7 to use Gram matrices rather than covariance matrices: $$\mathcal{L}_{\textrm{FroSSL}} = \log(\frobeniussq{Z_1^T Z_1}) + \log(\frobeniussq{Z_2^T Z_2}) + \mathcal{L}_{\textrm{MSE}}(Z_1, Z_2)$$
     $$\quad \quad \quad= \log(\frobeniussq{Z_1 Z_1^T}) + \log(\frobeniussq{Z_2 Z_2^T}) + \mathcal{L}_{\textrm{MSE}}(Z_1, Z_2)$$

    Assuming that each embedding is normalized to have unit norm, then both Gram matrices have 1 in each diagonal element. Then the rest of the proof then follows similarly to Proposition 1.

\section{Minibatch Time for Larger Batch Size and Dimensionality}
\label{appendix:extreme-minibatch}

\begin{table}[]
\caption{Comparison of the time/space tradeoffs for SSL algorithms trained on STL-10. We used a batch size of N=1024 and dimensionality D=1024.  The advantages of the reduced time complexity of FroSSL are more apparent here than in Table \ref{tab:time-complexities} of the main text due to the larger batch size and dimensionality. We encourage the reader to constrast this with Table \ref{tab:time-complexities}.}
\scalebox{0.7}{
\begin{tabular}{r|ccc|ccc|ccc|ccc}
\hline
 & SimCLR & MoCo v2 & BYOL & VICReg & Barlow & CorInfoMax  & \multicolumn{3}{c|}{MMCR} & \multicolumn{3}{c}{FroSSL (ours)} \\ 
 Loss Time Complexity          & \multicolumn{3}{c|}{$O(V^2 N^2)$} & $O(V D^2)$   & $O(V^2 D^2)$ & $O(V D^3)$  &\multicolumn{3}{c|}{$O(\min(D, N)^3)$} & \multicolumn{3}{c}{$O(V \min(D, N)^2)$} \\ \hline
Num. Views & 2 & 2 & 2 & 2 & 2 & 2 & 2 & 4 & 8 & 2 & 4 & 8 \\
VRAM Space (GB) & 5.5 & 6.7 & 6.3 & 5.4 & 5.4 & 5.5 & 5.4 & 10.4 & 19.6 & 5.4 & 10.2 & 19.6 \\
Minibatch Wall-time (ms) & 165 & 220 & 210 & 168 & 165 & 231 & 252 & 400 & 700 & 168 & 318 & 601 \\ \hline
\end{tabular}}
\label{tab:times-bigbatchsize}
\end{table}

\end{document}